\documentclass{llncs}
\usepackage{amsmath}
\usepackage{todonotes}
\usepackage{lscape}
\usepackage{amssymb}
\usepackage{amsfonts}
\usepackage{stmaryrd}
\usepackage{array}
\usepackage{listings}
\usepackage[caption=false]{subfig}
\usepackage[verbose]{placeins}
\usepackage[caption=false]{subfig}
\usepackage{tikz}
\usetikzlibrary{shapes, positioning, calc, chains}
\tikzset{myellipse/.style={ellipse,draw,minimum width=1cm,minimum height=1cm},
	myrectangle/.style={draw, rounded corners=10pt, minimum size=1.25cm},
	zeroarrow/.style = {-stealth,dashed},
	onearrow/.style = {-stealth,solid},
	c/.style = {circle,draw,solid,minimum width=2em,
		minimum height=2em},
	r/.style = {rectangle,draw,solid,minimum width=2em,
		minimum height=2em},
	b/.style = {rectangle,draw,solid,minimum width=2em,
		minimum height=3em}}
\begin{document}

\definecolor{numbergray}{gray}{0.5}

\lstdefinestyle{mgl}{
   columns=fixed,
	breaklines=true,
	numbers=left,
	numberstyle=\tiny,
	stepnumber=1,
	numbersep=5pt,
   showspaces=false,
   showstringspaces=false,
   frame=tblr,
   tabsize=2,
   frame=shadowbox,
   columns=fixed,
   basicstyle=\ttfamily\scriptsize,
   rulesepcolor=\color[rgb]{0.6, 0.6, 0.6},
    keywordstyle=\color[rgb]{0.5,0,0.34}\textbf,
   stringstyle=\color{blue},
   commentstyle=\color[rgb]{0.4,0.7,0.4},
   backgroundcolor=\color[rgb]{0.97,0.97,0.97},
   morekeywords={node, edge, attr, container, as, package, nsURI, iconPath, abstract, import,extends,
	diagramExtension, graphModel, incomingEdges, outgoingEdges, for, prime,containableElements,
	parameters, names, types},
	morestring=[b]",
   showtabs=false,
	morecomment=[l]{//},
	morecomment=[s]{/*}{*/},
   literate={0}{{{\color{numbergray}0}}}{1}%
		{1}{{{\color{numbergray}1}}}{1}%
		{2}{{{\color{numbergray}2}}}{1}%
		{3}{{{\color{numbergray}3}}}{1}%
		{4}{{{\color{numbergray}4}}}{1}%
		{5}{{{\color{numbergray}5}}}{1}%
		{6}{{{\color{numbergray}6}}}{1}%
		{7}{{{\color{numbergray}7}}}{1}%
		{8}{{{\color{numbergray}8}}}{1}%
		{9}{{{\color{numbergray}9}}}{1}
}

\newcommand{\includemgl}[3]{\lstinputlisting[style=mgl]{#1}}

\lstdefinestyle{style}{
   columns=fixed,
	breaklines=true,
	numbers=left,
	numberstyle=\tiny,
	stepnumber=1,
	numbersep=5pt,
   showspaces=false,
   showstringspaces=false,
   frame=tblr,
   tabsize=2,
   frame=shadowbox,
   columns=fixed,
   basicstyle=\ttfamily\scriptsize,
   rulesepcolor=\color[rgb]{0.6, 0.6, 0.6},
   keywordstyle=\color[rgb]{0.5,0,0.4}\textbf, 
   stringstyle=\color{blue},
   commentstyle=\color[rgb]{0.4,0.7,0.4},
   backgroundcolor=\color[rgb]{0.97,0.97,0.97},
	morekeywords={nodeStyle, edgeStyle, rectangle, ellipse, roundedRectangle,
	text, polyline, size, corner, position, value, color, lineStyle, lineWidth,
	SOLID, DASH, DASHDOT, DASHDOTDOT, DOT, decorator, location, relativeToMid,
	points, appearance, appearanceProvider, background, foreground, relativeTo,
	ARROW, CIRCLE, TRIANGLE, CENTER, MIDDLE, @, extends, movable},
	morestring=[b]",
   showtabs=false,
	morecomment=[l]{//},
	morecomment=[s]{/*}{*/},
   literate={0}{{{\color{numbergray}0}}}{1}%
		{1}{{{\color{numbergray}1}}}{1}%
		{2}{{{\color{numbergray}2}}}{1}%
		{3}{{{\color{numbergray}3}}}{1}%
		{4}{{{\color{numbergray}4}}}{1}%
		{5}{{{\color{numbergray}5}}}{1}%
		{6}{{{\color{numbergray}6}}}{1}%
		{7}{{{\color{numbergray}7}}}{1}%
		{8}{{{\color{numbergray}8}}}{1}%
		{9}{{{\color{numbergray}9}}}{1}
}

\newcommand{\includestyle}[3]{\lstinputlisting[style=style]{#1}}

\lstdefinestyle{java}{
   language=Java,
   breaklines=true,
   numbers=left,
   numberstyle=\tiny,
   stepnumber=1,
   numbersep=5pt,
   showspaces=false,
   showstringspaces=false,
   frame=tblr,
   tabsize=2,
   frame=shadowbox,
   columns=fixed,
   basicstyle=\ttfamily\scriptsize,
   rulesepcolor=\color[rgb]{0.6, 0.6, 0.6},
   keywordstyle=\color[rgb]{0.5,0,0.4},
   stringstyle=\color{blue},
   commentstyle=\color[rgb]{0.4,0.7,0.4},
   backgroundcolor=\color[rgb]{0.97,0.97,0.97},
   showtabs=false
}

\newcommand{\includejava}[3]{\lstinputlisting[style=java, float=b, caption=#2, label=#3, captionpos=b]{#1}}

\lstdefinestyle{xtext}{
   columns=fixed,
        breaklines=true,
        numbers=left,
        numberstyle=\tiny,
        stepnumber=1,
        numbersep=5pt,
   showspaces=false,
   showstringspaces=false,
   frame=tblr,
   tabsize=2,
   frame=shadowbox,
   columns=fixed,
   basicstyle=\ttfamily\scriptsize,
   rulesepcolor=\color[rgb]{0.6, 0.6, 0.6},
   keywordstyle=\color[rgb]{0.5,0,0.34}\textbf,
   stringstyle=\color{blue},
   commentstyle=\color[rgb]{0.4,0.7,0.4},
   backgroundcolor=\color[rgb]{0.97,0.97,0.97},
   morekeywords={grammar, with, hidden, generate, as, import, returns, current, terminal, enum, STRING, ID},
   keywordstyle=[2]{\textbf},
   morecomment=[l]{//}, 
   morecomment=[s]{/*}{*/}, 
   morestring=[b]",
   morestring=[b]',
   tabsize=2
}

\newcommand{\includextext}[3]{\lstinputlistinforEachg[style=xtext, float=tb, caption=#2, label=#3, captionpos=b]{#1}}

\definecolor{extension}{RGB}{174,48,0}

\lstdefinestyle{xtend}{
	language=Java,
	breakatwhitespace=true,
   columns=fixed,
        breaklines=true,
        numbers=left,
        numberstyle=\tiny,
        stepnumber=1,
        numbersep=5pt,
   showspaces=false,
   showstringspaces=false,
   frame=tblr,
   tabsize=2,
   frame=shadowbox,
   columns=fixed,
   basicstyle=\ttfamily\scriptsize,
   rulesepcolor=\color[rgb]{0.6, 0.6, 0.6},
   keywordstyle=\color[rgb]{0.5,0,0.34}\textbf,
   stringstyle=\color{blue},
   commentstyle=\color[rgb]{0.4,0.7,0.4},
   backgroundcolor=\color[rgb]{0.97,0.97,0.97},
 morekeywords={override,cached,case,default,extension,false,import,JAVA,WORKFLOWSLOT,let,new,null,private,create,switch,this,true,reexport,around,if,then,else,context},
 keywordstyle=[2]{\textbf},
 morecomment=[l]{//}, 
 morecomment=[s]{/*}{*/}, 
 morestring=[b]",
 tabsize=2,
 literate={filter}{{\textcolor{extension}{filter}}}{5}
          {empty}{{\textcolor{extension}{empty}}}{4}
}

\newcommand{\includextend}[3]{\lstinputlisting[style=xtend, float=tb, caption=#2, label=#3, captionpos=b]{#1}}

\title{ADD-Lib: Decision Diagrams in Practice}
\author{
  Frederik Gossen \and
  Alnis Murtovi \and
  Philip Zweihoff \and 
  Bernhard Steffen}
\institute{TU Dortmund University, Germany\\
\email{\{firstname.lastname\}@tu-dortmund.de}}
\maketitle
\begin{abstract}
  In the paper, we present the ADD-Lib, our efficient and easy to use framework for Algebraic Decision Diagrams (ADDs).
  The focus of the ADD-Lib is not so much on its efficient implementation of individual operations, which are taken by other established ADD frameworks, but its ease and flexibility, which arise at two levels: 
  the level of individual ADD-tools, which come with a dedicated user-friendly web-based graphical user interface, and at the meta level, where such tools are specified. 
  Both levels are described in the paper:
  the meta level by explaining how we can construct an ADD-tool tailored for Random Forest refinement and evaluation, and the accordingly generated Web-based domain-specific tool, which we also provide as an artifact for cooperative experimentation. In particular, the artifact allows readers to combine a given Random Forest with their own ADDs regarded as expert knowledge and to experience the corresponding effect.

\end{abstract}
\section{Introduction}
\label{sec:intro}
Taking decisions on the basis of some scenario profile is omnipresent: the evaluation of program conditions falls into this category, as well as typical (AI-based) classification approaches, like e.g., letter recognition, but also (recommender system-based) offer selection business decisions.
Algebraic Decision Diagrams provide a powerful framework for decision taking that can easily be adapted to a wide range of application scenarios just by 'playing' with the underlying algebraic structure to support e.g., deterministic, probabilistic and fuzzy decision paradigms.
Random Forests provide an interesting application scenario, which is not only very popular in practice, but which can particularly benefit from the ADD technology. 
In fact, in \cite{forests} we were able to improve the Random Forest evaluation by multiple order of magnitude.

State-of-the Art ADD frameworks like CUDD are a good basis to efficiently implement corresponding decision support systems, but require low level programming and dedicated extensions, e.g. to change the underlying algebraic structure.

In the paper, we present the ADD-Lib, our efficient and easy to use algebraic framework for dealing with decision diagrams.
Characteristic for ADD-Lib are not so much the efficient implementation of individual operations, which are taken from other established ADD frameworks, but its ease and flexibility, which arise at two levels:
\begin{itemize}
	\item
	At the level of individual ADD-tools, which come with a dedicated user-friendly Web-based graphical user interface supporting cooperative work.
	These tools are designed to	address the application expert and to not require any programming expertise.
	\item At the meta level, where such tools are specified. This comprises 
	the definition of a
	corresponding domain-specific language in term of its abstract and 
	concrete (graphical)
	syntax, the required algebraic structure, as well as intended GUI features.
\end{itemize}
Both levels are described in the paper: the meta level by explaining how we can construct an ADD-tool tailored for Random Forest refinement and evaluation, and the accordingly generated Web-based domain-specific tool, which we also provide as an artifact for cooperative experimentation.
In particular, the artifact allows readers to refine given Random Forests with their own ADDs expressing their dedicated domain knowledge and to experience the corresponding effect.

In Section~2, we recap various forms of decision diagrams and point the reader to relevant literature. 
After introducing the ADD-Lib, our framework for Algebraic Decision Diagrams in Section~3, we present a case study in which we augment a machine-learned decision model with expert knowledge in Section~4 conclude in Section~5. 

\section{Decision Diagrams}
The most widely known example of decision diagrams are (Reduced Ordered) Binary Decision Diagrams (BDDs)~\cite{Bryant1986}. 
For more than 30 years now, this data structure is state of the art for the representation of Boolean functions. 
Many other forms of decision diagrams~\cite{Bryant2018,Minato1993} have emerged since: 
reduced, ordered, binary, n-ary, those over Boolean values and others with more than just two terminals. 
In this section, we will give a brief overview of some variants and point to relevant literature. 

\begin{figure}
	\centering
	\subfloat[BDD for the Boolean function $(x_1~\land~x_2)~\lor~x_3$.]{
		\begin{tikzpicture}[x=5em, y=5em]
	\node[c] (x1) at (0, 0) {$x_1$};
	\node[c] (x2) at (1, -1) {$x_2$};
	\node[c] (x3l) at (0, -2) {$x_3$};
	\node[r] (0) at (0, -3) {$0$};
	\node[r] (1) at (1, -3) {$1$};
	\draw[onearrow]  (x1) -- (x2);
	\draw[zeroarrow] (x1) -- (x3l);
	\draw[zeroarrow] (x2) -- (x3l);
	\draw[onearrow]  (x2) -- (1);
	\draw[onearrow]  (x3l) -- (1);
	\draw[zeroarrow] (x3l) -- (0);
\end{tikzpicture}
		\label{fig:bdd_example}
	}%
	\hfil
	\subfloat[ADD for the function $(x_1~*~x_2)~+~x_3$.]{
		\begin{tikzpicture}[x=5em, y=5em]
	\node[c] (x1) at (0, 0) {$x_1$};
	\node[c] (x2) at (1, -1) {$x_2$};
	\node[c] (x3l) at (0, -2) {$x_3$};
	\node[c] (x3r) at (1, -2) {$x_3$};
	\node[r] (0) at (0, -3) {$0$};
	\node[r] (1) at (1, -3) {$1$};
	\node[r] (2) at (2, -3) {$2$};
	\draw[onearrow]  (x1) -- (x2);
	\draw[zeroarrow] (x1) -- (x3l);
	\draw[zeroarrow] (x2) -- (x3l);
	\draw[onearrow]  (x2) -- (x3r);
	\draw[onearrow]  (x3l) -- (1);
	\draw[zeroarrow] (x3l) -- (0);
	\draw[onearrow]  (x3r) -- (2);
	\draw[zeroarrow] (x3r) -- (1);
\end{tikzpicture}
		\label{fig:add_example}
	}
	\caption{Exemplary BDD and ADD representing analogous functions.}
	\label{fig:dd_example}
\end{figure}
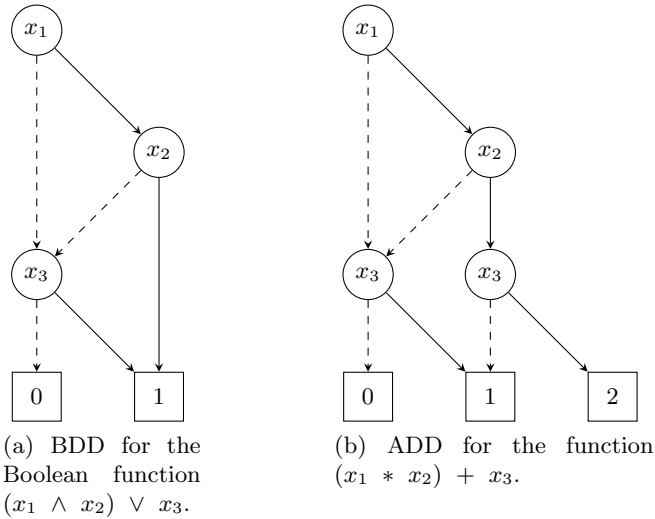
\textbf{Binary Decision Diagrams}, although the most popular form, were not the first kind of decision diagram. 
In fact, the first form was called a Binary-decision Program~\cite{Lee1959}, a representation that relies on the idea of many subsequent binary decisions until the result of the program is reached. 
Although it was not called one, this representation already constitutes a directed acyclic graph. 

The basic idea evolved to a graph-based data structure for Boolean functions~\cite{Akers1978}. 
With the elimination of duplicate and redundant nodes, the representation could be simplified, similar to the simplification of a Boolean formula. 

The breakthrough, however, was the introduction of an order on the Boolean variables of the input domain. 
For a given variable order, (Reduced Ordered) Binary Decision Diagrams (BDDs)~\cite{Bryant1986} are 
\begin{itemize}
  \item a canonical representation for Boolean functions ($\mathbb{B}^n \rightarrow \mathbb{B}$), 
  \item minimal in size, and 
  \item every input variable is encountered at most once per path from the root to a terminal node. 
\end{itemize}
For brevity, we will assume all decision diagrams to be reduced and ordered in what follows. 

Because BDDs represent Boolean functions, all logical operations, e.g. conjunction, disjunction, and negation, are naturally defined on them and efficient algorithms exist.
In this way, BDDs form an algebraic structure analogous to the standard Boolean logic. 
Binary Decision Diagrams are used for symbolic model checking \cite{bdd_sym_mc}, equivalence checking \cite{bdd_equiv}, VLSI-synthesis \cite{bdd_vlsi} and various other areas.
An exemplary BDD for the Boolean function $x_1 \wedge x_2 \vee x_3$ is displayed in Figure~\ref{fig:bdd_example}.

\textbf{Algebraic Decision Diagrams} (ADDs)~\cite{Bahar1993} are a generalization of BDDs that is of particular interest to our contributions. 
Rather than representing BDDs with a Boolean co-domain ($\mathbb{B}^n \rightarrow \mathbb{B}$), they are able to represent functions with an arbitrary co-domain ($\mathbb{B}^n \rightarrow A$).
At the same time, ADDs inherit their key properties, canonicity, minimality, and variable appearances, from BDDs. 

ADDs have their origin in \emph{Multi Terminal BDDs}~\cite{Fujita1997}, however, their association with an algebraic structure is the essential conceptual extension~\cite{Bahar1993}.
With that, we can define any algebraic structure, e.g. $(A, \circ)$, directly on the co-domain $A$. 
Every operation on this carrier set can be lifted to the level of their respective decision diagrams.

Similarly to the realization of standard Boolean operations on BDDs, efficient algorithms are known for ADDs, too.
Rather than conjunction, disjunction, and negation, we can, e.g., define a join operation $\circ$.
The operation can then be applied to (i)~elements of the co-domain as well as to (ii)~the corresponding ADDs.
For brevity, we denote the operations on the co-domains and those on the ADDs with the same symbols.

Use cases for ADDs include stochastic planning~\cite{Hoey1999}, stochastic model checking~\cite{Kwiatkowska2002}, and logic synthesis, verification, and testing of digital circuits~\cite{Bahar1993}.

An exemplary ADD with the associated algebraic structure $(\mathbb{Z}, +, *)$ is shown in Figure~\ref{fig:add_example}.
The depicted function $x_1 * x_2 + x_3$ is similar to that of previously seen BDD in the sense that multiplication replaces conjunction and addition replaces disjunction (cf. Figure~\ref{fig:bdd_example}).

\begin{figure}[t]
	\includegraphics[width=\textwidth]{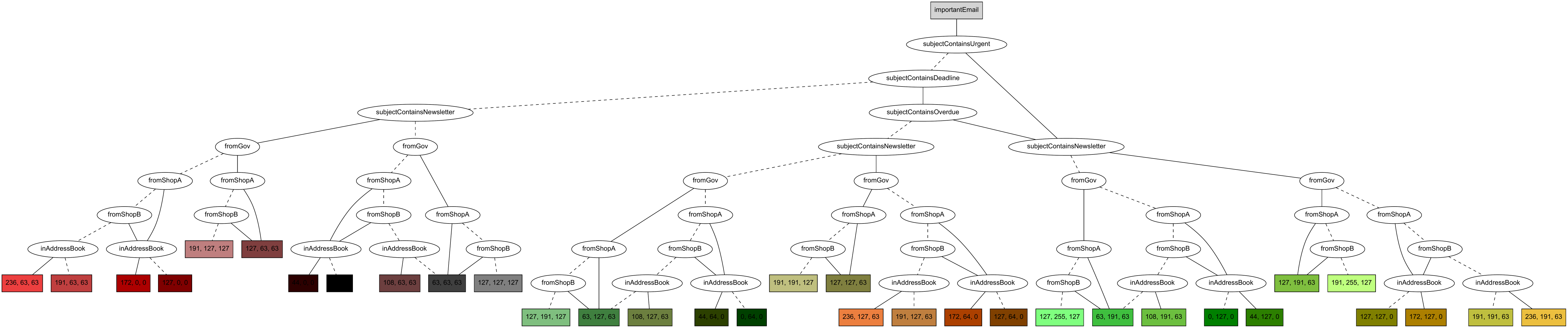}
	\centering
	\caption{Exemplary ADD with the RGB color space as its algebraic structure.}
	\label{fig:add_example_colours}
\end{figure}
ADDs are by no means limited to integers, or even numbers. 
A more complex example is shown in Figure~\ref{fig:add_example_colours} where the algebraic structure is defined over colors in the RGB space $[0 \ldots 255]^3$.
This example originates from an email classification service that assigns colors to emails based on some importance criterion~\cite{stressLDE}.
It is shown here solely to illustrate the flexibility of ADDs.

\section{ADD-Lib}
\label{sec:addlib}
There exist many implementations of the standard algorithms for Binary Decision Diagrams (BDDs), Algebraic Decision Diagrams (ADDs), and even for Zero-suppressed Decision Diagrams (ZDDs)~\cite{Brace1990,Lv2013,Somenzi2001}.
However, these realizations lack flexibility at their core: 
The underlying algebraic structures, e.g. the standard Boolean logic, are hard coded -- even for \emph{Algebraic} Decision Diagrams. 

In CUDD~\cite{Somenzi2001}, e.g., ADDs are limited to real numbers\footnote{Real numbers are represented by \texttt{double} values.} and the standard arithmetic operations. 
The framework is, in fact, so rigid that a change of the algebraic structure imposes a non-trivial adaptation of the library's core -- a change that would effect major parts of the code base. 

With the ADD-Lib~\cite{addlib}, we provide a framework that overcomes these limitations: 
This decision diagram framework is highly flexible and puts emphasis on the interchangeability of the underlying algebraic structure. 
At the same time, the implementation delegates computationally expensive operations to the well-established and robust C implementation of CUDD in a service-oriented fashion. 
In this way, the ADD-Lib inherits its broad range of functions and provides an easy-to-use yet flexible interface for the most common use cases. 

The definition of a new algebraic structure demands no more than a specification of its carrier set and the associated operations. 
This way, the algebraic structure can be defined directly on the co-domain of the ADDs. 
That very definition is then plugged into the ADD algorithms for the operations on the actual data structure. 
The user defines the simple underlying algebraic structure and gets the corresponding algebraic structure of ADDs with no additional effort. 

To facilitate the implementation of custom algebraic structures, the ADD-Lib comes with templates for the most common use cases. 
The categorization into 
\begin{itemize}
  \item groups and group-like structures, 
  \item rings and ring-like structures, and 
  \item various lattices and logics
\end{itemize}
allows to determine the required operations at compile time.
For example, a Boolean lattice will require conjunction, disjunction, and negation as well as a distinguished $0$ and $1$ element.

Consider an application that makes Boolean decisions based on some predicates, e.g. an email filter~\cite{stressLDE,lncs10000}. 
A standard BDD with its Boolean operations would be a suitable data structure to represent such a decision function. 
However, when some notion of certainty is required the standard Boolean logic $(\mathbb{B}, \land, \lor, \neg)$ is no longer an appropriate model. 
Instead, a three-valued logic or one of the many fuzzy logics is then more suitable to model the problem. 
A concrete instance of such a logic is the probabilistic fuzzy logic: 
\begin{align*}
  A_f := ([0, 1], \land_f, \lor_f, \neg_f) 
  \quad 
  \textrm{with} 
  \quad
  a \land_f b &:= a * b \\
  a \lor_f b &:= 1 - (1 - a) * (1 - b) \\
  \neg_f &:= 1 - a
\end{align*}

This custom definition of a logic is naturally defined on its carrier set, the interval $[0, 1]$. 
The ADD-Lib lifts this definition to ADDs and, with that, provides a data structure to model decision functions of the form $\mathbb{B}^n \rightarrow [0, 1]$. 
All operations that were defined on the carrier set, i.e. conjunction, disjunction, and negation, are then also applicable to these decision diagrams. 

Besides being an extremely flexible framework for ADDs, the ADD-Lib comes with powerful tools that facilitate experimentation with various forms of decision diagrams. 
It includes tools to visualize the in-memory data structure and to export it in common formats, including image formats and dot code. 

The many code generators that come with the ADD-Lib allow for its use in optimizing compilers or code generators~\cite{miaamics,stressLDE,forests,lncs10000}. 
ADDs are used as the intermediate representation in these applications, taking full advantage of their optimization potential. 
From that optimized, and in some cases optimal, representation an implementation in many common general-purpose programming languages can be generated fully automatically. 
Among the supported languages are Java, JavaScript, C, C++. 
The generated code implements the represented function in the form of a goto-program -- a hardly readable but very efficient implementation. 
This code can then be compiled together with its user application and allows for rapid evaluation of the ADD's decision function. 

\subsection{ADDs for Program Optimization}
An easily interchangeable algebraic structure allows for the use of ADDs in many new domains. 
We have used them primarily as an intermediate representation (IR) for program optimization. 
This concerns:  
\begin{itemize}
  \item \textbf{Domain-specific languages} that focus on a particular domain, e.g. Random Forests~\cite{forests} or decision rules~\cite{miaamics}.
  Constraining a language to one domain allows not only for a mindset-oriented design of the language~\cite{lncs10000} but also for a domain-specific optimization of the programs. 
  For languages that model decision services based on some predicates, ADDs with a domain-specific algebraic structure serve a suitable intermediate representation with desirable properties:
  They (i)~optimize decision functions, already as part of the standard algorithms, and they are (ii)~compositional, a property that is desirable also in the input language. 
  \item \textbf{General-purpose languages}. 
  With the right algebraic structure, we were able to optimize and to compile programs of the exemplary while-language~\cite{addcompiler}. 
  In fact, this radically new approach allowed us to significantly optimize an iterative Fibonacci implementation -- a program that generally appears to lack optimization potential.  
\end{itemize}

In this paper, we continue with this line of work and we use ADDs as an intermediate representation for a domain-specific graphical language that allows us to combine the power of machine learning with expert knowledge. 

\section{Augmenting Random Forests with Expert Knowledge}
Random Forests are one of the most widely known classifiers in machine learning. 
With Algebraic Decision Diagrams (ADDs) over three different algebraic structures, we were able to transform the original model into a single aggregated ADD in a semantics-preserving fashion. 
With this approach, it is possible to reduce the model's running time and also its size quite drastically~\cite{forests}. 

Being semantics-preserving, this approach fully inherits the forest's behavior:
where the forest makes the right decision, we make the right decision and vice versa. 
Machine learning methods in general, and Random Forests in particular, derive their model from a (representative) training dataset and infer some generalized decision model from it.
For new instances, this model is able to make a decision and, in this sense, generalizes its training dataset.  

Heuristic approaches like these are often very successful for problems that are insufficiently understood by humans. 
While good solutions can be found, these approaches typically yield little to no insight about the underlying phenomenon. 
At the same time no guarantees are given and a machine learned model remains purely heuristic. 

Expert knowledge, on the other hand, can only help if a problem is sufficiently understood. 
Where this is possible, however, it can provide hard guarantees about the resulting behavior. 

The question arises how to combine the advantages of both worlds. 
A method is needed to compose information form the two sources. 
As we have demonstrated in previous work, ADDs are a suitable data structure for Random Forests~\cite{forests} and also for domain-specific decision services~\cite{miaamics,lncs10000}. 
It is only natural to build upon this and to allow for their composition in yet another domain-specific language. 

At first, this creation of many graphical domain-specific languages -- sometimes for a single use case -- may appear overly expensive. 
However, with state of the art language development tools~\cite{xtext,antlr} and meta-modeling frameworks~\cite{cinco,pyro} this development has become much cheaper than anticipated.

\subsection{Metamodel-based Generation of Collaborative Domain-specific Languages}
Domain-specific languages aim at closing the gap between domain knowledge and technical requirements. 
Graphical domain-specific languages (DSLs), in combination with sophisticated mindset-supporting Development Environments \linebreak (mIDE)~\cite{lncs10000}, have proven to be particularly suitable because of their intuitive usability in many fields.

In practice, three major bottlenecks aggravate the introduction of mIDEs:
\begin{itemize}
	\item the enormous development effort for the required domain-specific languages and their enclosing mIDE, 
	\item the continuous delivery of enhanced versions of the DSL as a result of language evolution, and 
	\item the circuitous manual installation of the graphical modeling tools on each device and the synchronization between multiple users during collaboration.
\end{itemize}

However, with the right tools, a meta model-based generation of modeling tools can meet these challenges. 
The Cinco SCCE Meta Tooling Suite~\cite{cinco} is one example that has been designed to overcome the bottleneck of expensive tool development by providing a holistic, simplicity-driven~\cite{MarSte2010} approach for the creation of such domain-specific graphical mIDEs.
The main concept of Cinco is the generation of an entire mIDE from a high-level specification of the defined model structures, functionalities, and the user interface. 
The semantics of the specified modeling language is defined by means of code generators and model transformations~\cite{LyKoSt2018}.

Besides the natively supported generation of Eclipse-based mIDEs, Cinco supports the web as an alternative target platform. 
This is realized in Pyro~\cite{pyro,lybecait2018tutorial}, a tool that uses the very same specification to generate an mIDE for the web that is, in principle, analogous to the Eclipse variant. 

A web-based mIDE provides distributed access to a centralized instance of a model that can be used and edited by multiple users. 
The generated tool is easily accessible through a web browser and requires no installation or maintenance by its users. 
At the same time the distribution of new, enhanced versions of a DSL becomes trivial: a server update is immediately available to all users. 

Conceptually, Pyro is geared to modern online editors for collaborative work, like Google Docs, Microsoft Office 365, or solutions like ShareLaTeX/Overleaf. 

In this way, Cinco and Pyro, together, address the three barriers that prevent the wide use of DSLs:
Development becomes cheap, language evolution and distribution become trivial, and the mIDE could hardly be more accessible to its users. 

\subsection{A Graphical Language for Algebraic Decision Diagrams}
\label{subsec:lang}
Our graphical language for Algebraic Decision Diagrams relies on an algebraic structure, tailored to the aggregation of Random Forests~\cite{forests}.
At the same time, we aim to conform to the mindset of an expert and to allow for manual adaptations of the resulting decision model.
The goal is the creation of a sophisticated tool for the machine learning domain with relatively little effort.

Random Forests~\cite{Breiman2001,Ho1995} are a widely-known classifier in machine learning.
They are easy to understand, implement, and at the same time very successful in many domains.
The ensemble consists of a multitude of decision trees, which are constructed from a training dataset.
Each decision tree evaluates different predicates based on given features and results in leaf nodes representing the voted category.
The task of classification is realized by summing up the votes for each tree in the forest and deciding for the most frequent one.
In contrast to decision trees, Random Forests do not overfit on their training dataset~\cite{Breiman2001}.

However, being a purely heuristic method, no guarantees about the decision model's behavior can be given. 
To address this issue, we will improve the classification prediction through the composition with other manually created decision models, i.e. ADDs.
This composition of Random Forests with expert knowledge will allow to enforce at least some guarantees and, in this sense, regain control over the prediction behavior. 

Our domain-specific language builds upon the experience in~\cite{stressLDE,lncs10000}.
The modeling languages are threefold:
\begin{itemize}
  \item one model for the declaration of features and categories (also called classes), 
  \item multiple ADD models that, based on predicates over the features, determine one of the categories as the most likely outcome, and 
  \item one composition model that defines the aggregation of the individual ADDs.
\end{itemize}
The underlying algebraic structure of the regarded ADDs is defined as
\begin{align*}
  A &:= (\mathbb{R}^n, +, *, norm) 
\end{align*}
where the co-domain $\mathbb{R}^n$ is an $n$-dimensional vector space and addition $+$ and multiplication $*$ are defined component-wise.
In addition, we define a normalization on the vectors $\mathbf{v} = v_1, v_2, \ldots, v_n$ as 
\begin{align*}
  norm(\mathbf{v}) &:= \frac{1}{\sum_{i = 1}^{n} v_i} ~ \mathbf{v} \textrm{.}
\end{align*}

In the context of Random Forests, the aim is to model the frequencies, or weights, of the possible outcomes, i.e. the categories. 
Every component of a vector $\mathbf{v} \in \mathbb{R}^n$ represents the weight of one of the categories.

\subsubsection{Declaration of Features and Categories}
First, to declare the available features and categories, a dedicated \emph{Declaration} language is realized.
This language comprises corresponding node types for \emph{Feature} and \emph{Category}. 
The definitions will be referenced in the following languages and serve the purpose of seamless reuse (see Figure \ref{img:declaration} for an example).

\subsubsection{Algebraic Decision Diagrams}
Next, we present our \emph{DecisionDiagram} language (see Figure \ref{img:learned_trees} and \ref{img:manual_tree} for examples.). 
The abstract syntax is analogous to the structure of ADDs and consists of node types for \emph{Predicate}, \emph{Result} and a root node type called \emph{Function}. 
Moreover, the edge types \emph{TrueBranch} and \emph{FalseBranch} are used to connect the nodes.
\begin{itemize}
	\item The \emph{Predicate} node type corresponds to the inner nodes of ADDs. 
  Rather than limiting ourselves to Boolean variables, we allow for suitable predicates that compare an input feature against a constant real value.
	A predicate node type refers to a \emph{Feature} node type of the \emph{Declaration} language.
	The value against which the value of the feature is compared, is specified by an additional attribute of the \emph{Predicate} node.
	\item The \emph{Result} node type represents the ADD's terminal node.
	As our co-domain is defined as $\mathbb{R}^n$ and each entry of the $n$-dimensional vector corresponds to a category's associated weight, the \emph{Result} node type contains a list of nodes of type \emph{Value} which references a \emph{Category} from the \emph{Declaration} language.
	The weight for a \emph{Category} can be set as an attribute of \emph{Value}.
	\item The \emph{TrueBranch} and \emph{FalseBranch} edge types are used to connect \emph{Predicate} nodes to others of the same kind and also to the \emph{Result} nodes.
	By the use of connection constraints, a \emph{Predicate} node has exactly one \emph{TrueBranch}, one \emph{FalseBranch} successor, and at least one incoming edge.
	\emph{Result} node type, on the other hand, allow no outgoing edges, similar to the terminals of an ADD.
	\item The \emph{Function} node type points to the root of an ADD and assigns a name to it which enables the possibility of referencing ADDs and composing them as defined in the following \emph{Calculation} language.
\end{itemize}

\subsubsection{Composition of Algebraic Decision Diagrams}
The composition of multiple algebraic decision diagrams is realized by the \emph{Calculation} language combining references to decision diagrams and mathematical operations (see Figure \ref{img:composition} for an example).
The \emph{Calculation} language contains a \emph{DecisionDiagram} node type which holds a reference to a \emph{Function} node of the \emph{DecisionDiagram} MGL.
In addition, different calculation node types are defined to realize mathematical operations like \emph{Addition}, \emph{Subtraction}, \emph{Division} and \emph{Multiplication}.
To differentiate between associative and none associative operators, two corresponding abstract node types are specified supporting dedicated edge connection constraints.
Associative operator sub node types have two incoming \emph{SingleParameter} edges, whereas none associative sub node types have one \emph{LeftParameter} and one \emph{RightParameter} incoming edge.
This restriction will be enforced by generated validators in the running tool.

It is possible to generate code from the graphical models, e.g.\ to utilize composed ADDs for the evaluation of input data or to visualize the composed ADDs.
This transformation is realized by first creating ADDs with the ADD-Lib from the graphical models.
As the ADD-Lib already provides numerous code generators for different target languages, these can simply be invoked.

\subsubsection{Realization in Cinco and Pyro}
The abstract syntax of our three graphical languages is defined with the Meta Graph Language (MGL) of Cinco.
The MGL for the \emph{DecisionDiagram} language is displayed in Listing \ref{lst:tree} which highlights how the previous node and edge types as well as their attributes are defined.
As the \emph{Predicate} node type references a node of type \emph{Feature} and the \emph{Value} node type references a node of type \emph{Category}, the \emph{Declaration} MGL has to be imported (see Line $1$, $9$ and $19$).
Already in the MGL static semantics such as edge constraints can be specified where Cinco automatically generates corresponding validations in the modeling environment.
\emph{Result} is of type container as it can contain multiple nodes of type \emph{Value}.

\begin{figure}[t]
	\includemgl{tree.mgl}{{tree.mgl} for the AADL project}{lst:mgl}
	\caption{Abstract syntax of the DSL for decision diagrams (DecisionDiagram.mgl).}
	\label{lst:tree}
\end{figure}

From the short description of the graph syntax, we are able to generate a mindset-supporting Development Environment (mIDE)~\cite{lncs10000} fully automatically.
This generative approach is the crucial enabler for the development of domain-specific languages. 
Only with their development being as easy and cheap as this, domain-specific languages as focused as the ones discussed in this paper can eventually pay off.

\subsection{Introducing Expert Knowledge with our Web mIDE}
\label{sec:tutorial}
%
%
In this section, we use the newly created domain-specific language (Sec.~\ref{subsec:lang}) for ADDs to incorporate expert knowledge into a machine-learned model. 
The example of a Random Forest, that was trained on the popular Iris dataset~\cite{iris-dataset}, serves a subject throughout the remainder of this paper. 

The demonstration of our graphical language is structured as follows:
\begin{itemize}
	\item We first see how three learned decision trees from the Iris data set can be represented as ADDs within our graphical modeling tool.
	\item We will then create an ADD by hand which contains knowledge of a domain expert.
	\item Finally, we aggregate the three learned decision trees into one semantic preserving ADD and then aggregate this ADD with the ADD-Lib, which was created by hand into one final ADD.
	This final ADD represents a Random Forest representing knowledge from both the learned decision tree and the domain expert.
\end{itemize}

\subsubsection{Random Forest Transformation to ADD Models}
%

A Random Forest is trained on a dataset, the Iris dataset~\cite{iris-dataset} in this case. 
The dataset lists dimensions of Iris flowers' sepals and petals for three different species. 
As a result all predicates in the forest reason about these features in the sense that they compare them against a constant threshold. 
\begin{figure}[t]
	\centering
	\includegraphics[width=0.8\textwidth]{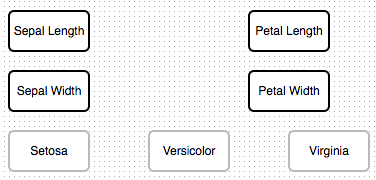}
	\caption{Iris flower declared features and categories.}
	\label{img:declaration}
\end{figure}
\FloatBarrier
The number of individual trees in a Random Forest is, in principle, not limited. 
For illustration purposes, we choose a small size of only three trees. 
Note, however, that neither the Random Forest nor our aggregation is limited to such a small number. 
In fact, we were able to aggregate extremely large forests with up to 10,000 trees~\cite{forests}.

With trees being structurally similar to decision diagrams of our language, their transformation is straightforward. 
In fact, decision trees constitute a subset of our modeling language.  

The Iris flower example declaration model and three learned trees can be created using the \emph{Create Iris} context menu action available on a \emph{Declaration} model instance in the running IDE.
The declaration model (see Fig. \ref{img:declaration}) contains instances of the prior defined \emph{Feature} and \emph{Category} node types.

\begin{figure}[t]
	\includegraphics[width=\textwidth]{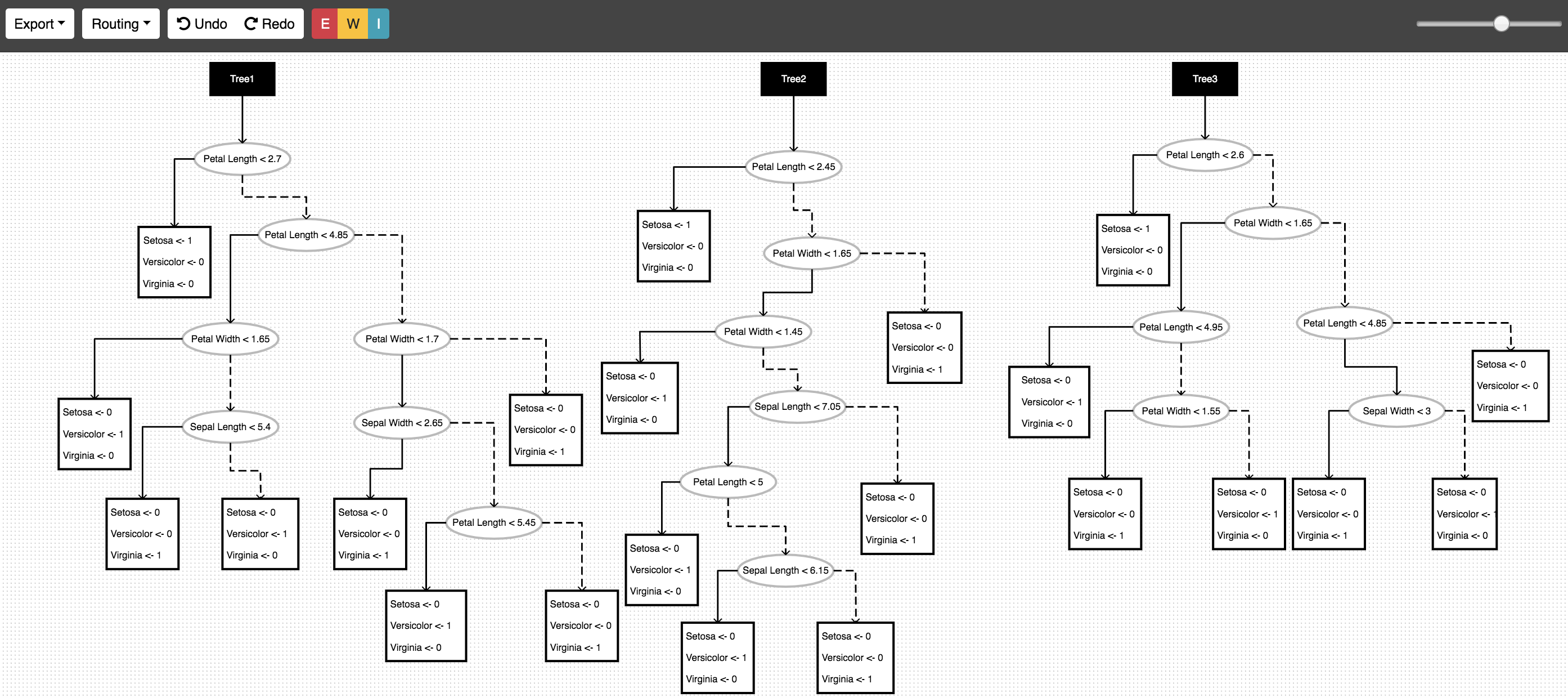}
	\caption{Learned and imported decision diagram}
	\label{img:learned_trees}
\end{figure}

The created feature and category nodes are used to define predicates and result nodes of the ADDs as instance of the \emph{Decision Diagram} DSL.
As a result of this, the imported ADDs (see Fig. \ref{img:learned_trees}), which have been learned and extracted from the Random Forest refer to the defined features and categories.
These three learned decision trees represent a Random Forest learned from the Iris flower data set.
They are similar in size and the leaf nodes contain exactly one class weight which is set to $1$ while all other class weights are set to $0$.
The class weight set to $1$ represents the predicted class of the original decision tree.

\subsubsection{Aggregation with Composition Models}
\begin{figure}[t]
	\includegraphics[width=\textwidth]{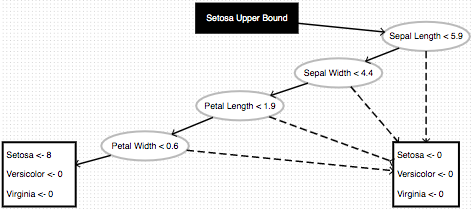}
	\caption{Manually created ADD enforces expert knowledge about Iris Setosa.}
	\label{img:manual_tree}
\end{figure}

As a next step, the \emph{DecisionDiagram} DSL can be used to create an additional decision diagram, based on the same feature declaration.
In contrast to the decision trees displayed in Figure~\ref{img:learned_trees}, the decision diagram in Figure~\ref{img:manual_tree} was not learned but created manually by a domain expert, i.e.\ a non-programmer with extensive knowledge of the domain, in this case the Iris flowers.

Assume, that this domain expert cares particularly about the Iris species \emph{Setosa}.
He or she may model, e.g. the decision diagram that has one path which leads to a leaf and ensures that the \emph{Setosa} flower is chosen as the predicted class.
For this reason the class weight of \emph{Setosa} is set to $8$ while the remaining class weights are set to $0$.
This single decision diagram in this case influences the predicted class in the same way $8$ decision trees would.
In all other cases the reached leaf contains class weights which are set to $0$ which can be interpreted as cases which we do not want to affect.

\begin{figure}[t]
	\includegraphics[width=\textwidth]{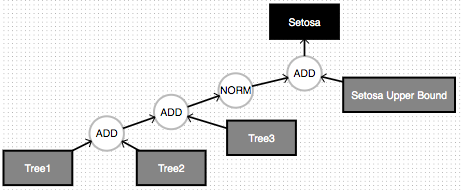}
	\caption{Composition of the three learned ADDs and the manually created ADD.}
	\label{img:composition}
\end{figure}

The learned decision trees in Figure \ref{img:learned_trees} and the manually created decision diagram in Figure~\ref{img:manual_tree} are now composed into one ADD in order to include the expert knowledge represented by the manually created ADD into the Random Forest.
This composition in Figure \ref{img:composition} can be divided into three parts.
First, the three learned decision trees are aggregated into one ADD which in essence represents the Random Forest consisting of these three single decision trees.
The aggregation is realized by the $+$ operation, i.e. component-wise addition of two class vectors, which is then lifted to the ADD.

Next, the class weights of the aggregated ADD are normalized by applying the monadic $norm$ operation (Sec.~\ref{subsec:lang}) which transforms the absolute class weights to relative values in the interval $[0, 1]$.

Finally, the normalization of the aggregated ADD is aggregated together with the manually created ADD resulting in the final ADD.
As the class weights of the aggregated ADD, resulting from the three learned ADDs, are normalized values between $0$ and $1$, the decision of the manually created ADD is weighted much stronger.
The class weight of the \emph{Setosa} category is set to $8$ in one of the leafs of the manually created ADD which ensures that if the features' input values are such that the path to the leaf is traversed, the predicted category in the final aggregated ADD will also be the \emph{Setosa} category.

\begin{figure}[t]
	\centering
	\includegraphics[width=\textwidth]{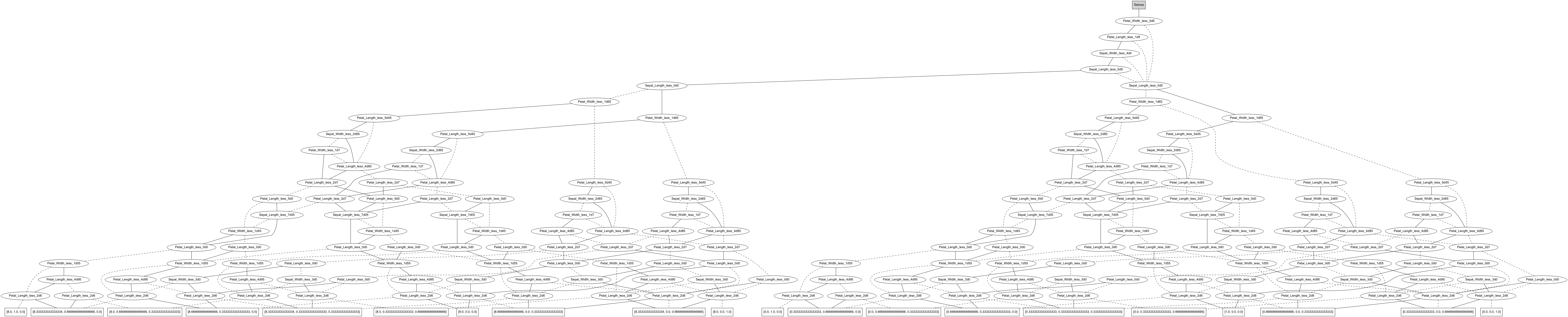}
	\caption{The final aggregated ADD.}
	\label{img:generated_dot}
\end{figure}

As mentioned, the ADD-Lib provides various code generation for different target languages.
An automatically generated visualization for the final aggregated ADD is displayed in Figure~\ref{img:generated_dot}.

The underlying code generators all rely on ADDs as their intermediate representation. 
Decision functions are automatically optimized by the properties of ADDs and allow for straightforward code generation. 
Only the transformation from models of the input language to the ADDs must be addressed. 
For the graphical languages in this example the transformation is straightforward along the acyclic structure of the models. 
All complexity is delegated to the ADD-Lib's core where well-studied and efficient ADD algorithms perform the transformation in a service-oriented fashion.

Besides generating code for visualization purposes, the code generators can also be used to generate an implementation of the represented function in various general-purpose programming languages. 
In our example, we use this to generate a JavaScript implementation fully automatically. 
The generated code can then be embedded in a web application that, in this case, provides a graphical interface for the evaluation of test data.

Given the four features sepal length, sepal width, petal length and petal width as input, the generated tool uses the ADD to predict the flower category.


\section{Conclusion}
\label{sec:concl}
We have presented the ADD-Lib, our efficient and easy to use  algebraic framework for dealing with decision diagrams.
Characteristic for ADD-Lib is its ease and flexibility, which arise at two levels: the level of individual 
ADD-tools, which come with a dedicated user-friendly Web-based graphical user interface, and at the 
meta level, where such tools are specified.
We have illustrated the  meta level by explaining how we can construct an ADD-tool tailored for Random Forest 
refinement and evaluation.
The according automatically generated Web-based domain-specific tool for Random Forest refinement and evaluation is provided as an artifact for readers to reply the development of Section 4, and to combine given Random Forests with their own ADDs regarded as expert knowledge and to experience the corresponding effect.

The kernel ADD-Lib, as well as the Cinco and the Pyro frameworks are open source but not yet presented
as a dedicated homogeneous artifact.
Thus readers could, in principle, also replay our meta-level development.
This may, however, be quite hard for newcomers.
We are therefore planning to develop a dedicated domain-specific meta tool for this purpose.

\bibliographystyle{splncs04}
\bibliography{literature}
\end{document}